# Hyperspectral Classification Based on Lightweight 3-D-CNN With Transfer Learning

Haokui Zhang, Ying Li, Yenan Jiang, Peng Wang, Qiang Shen, and Chunhua Shen

*Abstract*— Recently, hyperspectral image (HSI) classification approaches based on deep learning (DL) models have been proposed and shown promising performance. However, because of very limited available training samples and massive model parameters, DL methods may suffer from overfitting. In this paper, we propose an end-to-end 3-D lightweight convolutional neural network (CNN) (abbreviated as 3-D-LWNet) for limited samples-based HSI classification. Compared with conventional 3-D-CNN models, the proposed 3-D-LWNet has a deeper network structure, less parameters, and lower computation cost, resulting in better classification performance. To further alleviate the small sample problem, we also propose two transfer learning strategies: 1) cross-sensor strategy, in which we pretrain a 3-D model in the source HSI data sets containing a greater number of labeled samples and then transfer it to the target HSI data sets and 2) cross-modal strategy, in which we pretrain a 3-D model in the 2-D RGB image data sets containing a large number of samples and then transfer it to the target HSI data sets. In contrast to previous approaches, we do not impose restrictions over the source data sets, in which they do not have to be collected by the same sensors as the target data sets. Experiments on three public HSI data sets captured by different sensors demonstrate that our model achieves competitive performance for HSI classification compared to several state-of-the-art methods.

*Index Terms*— 3-D lightweight convolutional network (3-D-LWNet), deep learning (DL), hyperspectral classification, transfer learning.

## I. Introduction

HYPERSPECTRAL images (HSIs) typically contain abundant spectral and spatial information, offering a significant opportunity for land-cover classification. Rich spectral and spatial information is particularly beneficial to discriminate different objects of interest and also increases the dimensionality of samples that may affect the classification accuracy and efficiency [1]–[3]. In order to avoid this problem, effective and efficient feature extraction methods are necessary. However, HSIs are more complex than RGB images; therefore, performing HSI feature extraction is still a challenging task [4], [5].

In the early days of HSI classification, feature extraction focused only on spectral information. Approaches exploiting merely the spectral information fail to capture important spatial variability, generally resulting in poor performance. In fact, in HSIs, different objects may exhibit similar spectral features, whereas the same objects in different locations may emerge with different spectral features. For such objects, it is very difficult to classify with the use of spectral features alone.

To improve classification performance, recent studies have recommended combining spectral information with spatial information to extract spectral–spatial features. There are two main spectral–spatial feature extraction strategies. The first exploits the spectral and spatial contextual features separately [6], [7] and the second strategy works by fusing spatial information with spectral features to produce joint features [8]. For example, 3-D scattering wavelet filters [9] generated at different scales and frequencies have been applied on hyperspectral data to extract spectral–spatial features. Such a combination of spectral information and spatial information further improves the classification accuracy [6]–[9].

Most classification approaches are, however, based on handcrafted features and conventional learning models. First of all, handcrafted features are highly dependent on domain knowledge. Second, it is difficult to address the requirement of considering all the details embedded in all forms of real data using predesigned handcrafted features. To further improve the accuracy of HSI classification, more robust features and more powerful models are necessary.

Since 2012, when AlexNet [10] won the ImageNet classification challenge, deep learning (DL) has become a hot topic in computer vision including image classification [11], object detection [12], [13], tracking [14], and semantic segmentation [15]. One of the most significant advantages of DL is that it allows the extraction of efficient deep features from raw images in an end-to-end manner. Very recently, DL models have been introduced into HSI classification, leading to significant achievements [16]–[19]. For instance, Chen *et al.* [16], [17] first applied unsupervised deep feature learning, including stacked autoencoder (SAE) and deep belief network (DBN), for spectral–spatial feature extraction and

Manuscript received August 29, 2018; revised January 7, 2019; accepted February 25, 2019. This work was supported in part by the Foundation Project for Advanced Research Field of China under Grant 614023804016HK03002, in part by the National Natural Science Foundation of China under Grant 61871460 and Grant 61876152, in part by the National Key Research and Development Program of China under Grant 2016YFB0502502, and in part by the Innovation Foundation for Doctor Dissertation of Northwestern Polytechnical University under Grant CX201816. *(Corresponding author: Ying Li.)*

H. Zhang is with the Shaanxi Provincial Key Laboratory of Speech and Image Information Processing, School of Computer Science, Northwestern Polytechnical University, Xi'an 710129, China, and also with the School of Computer Science, The University of Adelaide, Adelaide, SA 5005, Australia (e-mail: hkzhang1991@mail.nwpu.edu.cn).

Y. Li, Y. Jiang, and P. Wang are with the Shaanxi Provincial Key Laboratory of Speech and Image Information Processing, School of Computer Science, Northwestern Polytechnical University, Xi'an 710129, China (e-mail: lybyp@nwpu.edu.cn; nyejiang@126.com; peng.wang@nwpu.edu.cn).

Q. Shen is with the Faculty of Business and Physical Sciences, Aberystwyth University, Aberystwyth SY23 3DB, U.K. (e-mail: qqs@aber.ac.uk).

C. Shen is with the School of Computer Science, The University of Adelaide, Adelaide, SA 5005, Australia (e-mail: chunhua.shen@adelaide.edu.au).

Color versions of one or more of the figures in this paper are available online at http://ieeexplore.ieee.org.

Digital Object Identifier 10.1109/TGRS.2019.2902568

classification. Supervised convolutional neural network (CNN) models, such as 2-D-CNN [18], 3-D-CNN [19], [20], and ResNet [21], [22], have been successfully exploited to extract deep spectral–spatial features and show the state-of-the-art performance.

Unlike natural image classification, HSI classification is a classification task involving 3-D data. As such, in HSI, the data structure is more complex but the number of labeled samples may be very limited. Almost all advanced CNN models such as ResNet [23] and MobileNet [24] are 2-D models, while the applications of which are focused on 2-D visual tasks. It is practically difficult to employ such 2-D models to perform 3-D HSI classification without any appropriate adjustments. Another problem is the number of available training samples. With the development of CNN techniques, the model scales (especially the depths ) increase rapidly. For instance, the depth of ResNet has increased to more than 1000 convolution layers. A large number of training samples are necessary in order to train such large scale networks. Without enough sufficient training samples, a very deep model that has a powerful representation capacity may suffer from overfitting.

For natural image classification, the number of labeled samples in the widely used data sets may vary from tens of thousands to tens of millions, such as ImageNet, VOC2007, VOC2012, and COCO data sets. For HSI classification, the number of available training samples in the commonly studied HSI data sets still varies from thousands to tens of thousands, such as Indian Pines and Pavia University scene data sets. Recent experimental results have shown that deep models generally perform better than shallow models. However, because of limited training samples, the CNN models employed in HSI classification typically consist of only less than five convolution layers. In other words, it is challenging to apply very deep CNN models to 3-D HSI classification with limited training samples, thereby restricting the achievement of the full potential of CNN models.

In this paper, we propose a deep 3-D lightweight convolutional network (3-D-LWNet) for HSI classification. Unlike conventional 3-D-CNN models in the HSI literature (see [19], [20]), which only use three 3-D convolution layers, the proposed 3-D-LWNet can employ tens of 3-D convolution layers. This development is more in line with the current trend of building DL models. In the meantime, the parameters involved within the 3-D-LWNet are much fewer than those of conventional 3-D-CNNs, which is more beneficial for problems with limited samples. In addition, to further alleviate the problem of HSI having limited training samples, we also adopt two transfer learning strategies in our framework: 1) pretraining a 3-D model in the HSI data sets that contain a relatively larger number of training samples and, subsequently, transferring it to suit the target HSI data sets and 2) pretraining a 3-D model in the natural image data sets that contain a large number of 2-D image samples and then transferring it to fit the target HSI data sets. We compare our framework with the aforementioned state-of-the-art CNN-based techniques on three real HSI data sets. Experimental results demonstrate that the proposed approach outperforms those conventional 3-D-CNN-based HSI classification methods.

The work of this paper focuses on employing deep 3-D-CNN to HSI classification under the condition of limited training samples. The proposed 3-D-LWNet is combined with transfer learning and achieves state-of-the-art performance in terms of classification accuracy. The main contributions of this paper are outlined as follows.

1) To the best of our knowledge, this is the first that 3-D-CNN consisting of tens of convolution layers is introduced into HSI classification.
2) The proposed 3-D-LWNet reduces the number of parameters required by the network for HSI classification with its parameters and computation cost being much less than those required by 3-D convolutional neural network and logistic regression (3-D-CNN-LR) (which represents the state-of-the-art 3-D-CNN-based HSI classification models). Interestingly, unlike natural RGB image classification where the reduction of network parameters and computation cost usually reduces the classification accuracy (e.g., MobileNet [24]), in our work, 3-D-LWNet greatly reduces network parameters and computation while improving classification accuracy.
3) In order to address the overfitting problem caused by limited training samples, transfer learning is adopted. Combined with 3-D-LWNet, two alternative transfer learning strategies are proposed: cross-sensor strategy and cross-modal strategy. With the former, we transfer 3-D-LWNet between different HSI data sets captured by the same sensor or different sensors. This forms a sharp contrast with the previous work that only transfers models between HSI data sets acquired by the same sensor, enabling model transfer between HSI data sets captured by different sensors for HSI classification, for the first time. The latter strategy has never been previously attempted, pretraining a 3-D-CNN on 2-D RGB natural image data sets and transferring it to suit 3-D HSI data sets through fine tuning, resulting in promising classification performance.

The remainder of this paper is organized as follows. Section II provides an introduction to the related work. Section III presents the details of our frameworks including the structure of 3-D-LWNet and the implementation details of transfer learning. We describe the data sets and experimental setups, discuss the experimental results, and empirically compare the proposed method with other 3-D-CNN-based HSI classification methods in Section IV. Finally, conclusions are presented in Section V with future work pointed out.

## II. Related Work

### A. DL for HSI Classification

Generally speaking, DL consists of four basic types of model, including SAE, DBN, CNN, and recurrent neural network (RNN). All four DL model types have found their applications in HSI classification literature.

An initial attempt can be found in [16], where Chen *et al*. adopted an SAE to extract spectral features and spatial features and then joint them to form spectral–spatial features. Spectral information does not require any preprocessing, but spatial information has to be flattened to a 1-D vector, as SAE can only handle 1-D input. Following this original work, the use of a DBN instead of SAE is reported in [17]. Similarly, Ma et al. [25] employed an SAE to learn effective features and added a relative distance prior in the subsequent fine-tuning process. Both SAE and DBN can extract deep features, but SAE and DBN cannot extract the spatial information efficiently because they need to flatten the spatial information into 1-D vectors, which does not retain the same spatial information that the original image may contain.

Compared to SAE and DBN, CNNs have been employed to HSI classification only recently. However, statistically, the number of papers regarding the use of CNNs for HSI classification grows fastest and the performance of CNNs is generally better. There are three main types of CNN: 1-D-CNN, 2-D-CNN, and 3-D-CNN. Running 1-D-CNN-based HSI classification, the kernels of a convolution layer convolve the input samples along the spectral dimension. Hu et al. [26] carried out HSI classification with 1-D-CNN containing four layers: one convolution layer followed by one pooling layer and two fully connected layers. Mei et al. [27] exploited a similar 1-D-CNN to classify HSI. For 2-D-CNN-based HSI classification approaches, HSIs are always compressed via a certain dimension reduction algorithm, such as principal component analysis (PCA) [28] and independent component analysis [29], and then convolved with 2-D kernels. Makantasis et al. [30] exploited randomized PCA to condense the spectral dimensionality of the entire HSI first, followed by applying a 2-D-CNN to extract deep features from the compressed HSI. In [31], the top three principal components are extracted from the raw HSI by the use of PCA, with the condensed HSI put through a 2-D-CNN to extract spatial features. As HSIs are 3-D data, it is reasonable to expand 2-D-CNN to 3-D-CNN for HSI classification. Both in [19] and [20], 3-D-CNNs are directly employed to learn deep spectral–spatial features. In particular, the former utilizes a large-scale 3-D-CNN that takes cubes of $27 \times 27$ in space size as input, whereas the latter uses a much more compact 3-D-CNN with input cubes of $5 \times 5$ in size. Zhong et al. [21] employed spectral and spatial residual blocks consecutively to learn spectral and spatial representations separately.

RNN is mainly designed to handle sequential data. HSIs can be seen as a set of orderly and continuing spectral sequences. Therefore, the RNN models have been recently introduced into HSI classification by analyzing HSI data in spectral sequences. Compared with HSI classification methods based on SAE, DBN, or CNN, approaches based on RNNs are relatively few. Mou et al. [32] attempted to use RNN to capture the sequential property of a pixel vector of hyperspectral data to perform HSI classification. Wu and Prasad [33] proposed a convolutional RNN for HSI classification, consisting of a few convolution layers followed by recurrent layers. As can be seen later, these approaches are all different from what we are proposing in this paper.

### B. CNN Architectures

Since AlexNet was proposed in 2012, a number of efficient DL models have been proposed. Among all these, four models GoogleNet [11], ResNet [23], DenseNet [34], and MobileNet [24] are related to our model proposed in the following, and they also show the development trends of DL, deeper in depth while lower in computation cost. The first one of these is the most basic of so-called Inception series, ResNet and DenseNet are famous for their extreme depth, and the last one is well known for its low computation cost.

*1) GoogleNet:* It consists of multiple inception modules, each of which contains four different convolution paths, and it is the most basic model of the Inception series [11]. Based on GoogleNet, Ioffe and Szegedy [35] proposed Inception-V1, which introduced batch normalization into inception modules to overcome internal covariate shifts. Batch normalization allows for the use of much higher learning rates and offers more flexibility regarding model initialization. As such, it almost has become a necessary layer for the network models proposed since 2015. Szegedy et al. [36] also further developed GoogleNet and proposed Inception-V2 and Inception-V3. In Inception-V2, they adopted batch normalization, factorization, and made other additional minor changes. Szegedy et al. [37] improved the previous Inception modules and proposed Inception-V4 and Inception-ResNet, where shortcut connections are also employed in Inception modules.

*2) ResNet:* It employs shortcut connections to overcome the degradation problem, where accuracy gets saturated and then degrades rapidly with the network depth increasing. In addition, in order to reduce the time complexity, He et al. [23] proposed a novel structure named "bottleneck." Based on shortcut connection and the newly introduced bottleneck layers, He et al. [23] increased the depth of the network to more than 1000 layers and obtained excellent performance in image classification. In addition, ResNet has also been used for object detection, such as the work of faster region-based convolutional neural network and you only look once.

*3) DenseNet:* It connects each layer to every other layer in a feedforward fashion. As with ResNet that builds the whole network by stacking several residual units, DenseNet consists of multiple dense blocks. In an $L$-layer dense block, there are $L(L+1)/2$ direct connections. Most recently, Wang et al. [52] have proposed a variant of DenseNet architecture called PeleeNet that follows the innovative connectivity pattern of DenseNet while adopting two-way dense layers to obtain different receptive fields.

*4) MobileNet:* It employs depthwise separable convolutions to reduce the computation in the network and applies pointwise convolutions to combine the features of separable channels. Based on MobileNet-V1, MobileNet-V2 was also proposed to employ inverted residuals and linear bottlenecks, leading to further improved performance [24], [38].

### C. Transfer Learning

DL models have already achieved significant successes in a range of fields, including classification, detection, tracking, and so on. However, many models work well only with a large volume of training samples. In particular, for classification and recognition, the success is based on both advanced models and a large number of available training samples. Lack of sufficient training samples may lead to a poor performance. In such cases, it would be helpful if transfer learning is adopted.

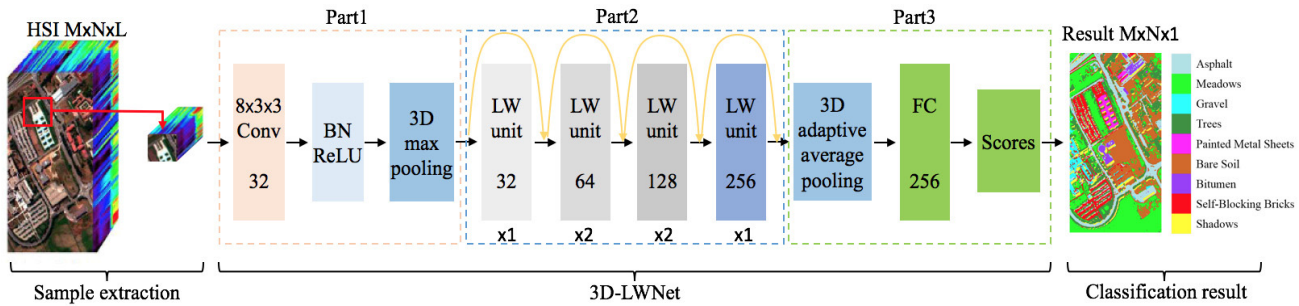

Fig. 1. 3-D-LWNet-based HSI classification framework. The first step is sample extraction, where $S \times S \times L$-sized sample is extracted from a neighborhood window centered around the target pixel. Once samples are extracted from raw HSI, they are put through the 3-D-LWNet to extract deep spectral–spatial features and to calculate classification scores.

Transfer learning focuses on storing knowledge gained while solving one problem and applying it to a different but related problem [39]. Broadly speaking, the goal of transfer learning is to use training data from related tasks to aid learning on a future problem of interest [40]. There are a large number of transfer learning strategies that have been proposed for different situations, including the learning of small sets of relevant features that are shared across a variety of tasks [41], [42]. In [41], when multiple classification tasks and different labeled data sets have a common input space, Jebara proposed a method to compute a common feature selection and configuration of kernel for multiple support vector machines trained on different yet interrelated data sets. Argyriou *et al.* [42] presented a method for learning a low-dimensional representation that is shared across a set of multiple related tasks. In the work on learning intermediate representations, Ando and Zhang [43] reported a general framework for learning predictive functional structures from multiple tasks. Raina *et al.* [44] proposed an algorithm for constructing the covariance matrix for an informative Gaussian prior. Based on boosted decision stumps, Torralba *et al.* [45] presented a multitask learning procedure to find common features that can be shared across the classes.

A common strategy of transfer learning is pretraining a model on one data set, which consists of a large number of labeled samples, such as ImageNet, and then transferring lightweight the pretrained model to the target data set to fine-tune. For data sets that contain very limited training samples, the use of transfer learning is extremely important, especially when the adopted model is deep CNN, which generally has a massive number of parameters. As HSI data sets always contain very limited training samples, transfer learning can play an important role. An early attempt to address this issue can be found in [46], where Yang *et al.* combined transfer learning with a two-branch CNN to learn the deep features from HSIs.

## III. PROPOSED METHOD

As indicated previously, DL models have been introduced into HSI classification and obtained good performance. However, HSI classification approaches based on DL still have room for improvement. The outstanding performance of ResNet proves that depth is very important for DL-based image processing methods. Inspired by this, we developed a very deep 3-D convolutional network for HSI classification. In this section, we give the details of the proposed network.

### A. 3-D-LWNet-Based Classification Framework

The framework of the HSI classification is shown in Fig. 1. It consists of three parts, including samples extraction, 3-D-LWNet, and classification result. The structure of HSI is 3-D, so it is intuitive to implement a 3-D model for classification. In sample extraction, we extract $S \times S \times L$-sized cube as a sample and each cube is extracted from a neighborhood window centered around a pixel. $S$ and $L$ are the spatial size and the number of spectral bands, respectively. The label of each sample is that of the pixel located in the center of this cube.

*1) 3-D-LWNet:* Once 3-D samples are extracted from HSI, we feed them into the 3-D-LWNet model that is itself composed of three parts to obtain the classification scores.

1) In part one, samples are grouped in batches of size $b$ (where each batch is $[b, 1, L, S, S]$-sized) and put through the first convolution layer, the batch normalization layer with rectified linear unit (ReLU) function, and 3-D max pooling layer. In the first convolution layer, the input batch is convolved with 32 $8 \times 3 \times 3$-sized 3-D kernels without padding. The output is a $b \times 32 \times (L-7) \times (S-2) \times (S-2)$-sized volume. After applying batch normalization and ReLU function, the $b \times 32 \times (L-7) \times (S-2) \times (S-2)$-sized volume is sent to the first 3-D max pooling layer with $3 \times 3 \times 3$-size kernel, stride of 2. The output of 3-D max pooling is $b \times 32 \times \lfloor (L-7)/2 \rfloor \times \lfloor (S-2)/2 \rfloor \times \lfloor (S-2)/2 \rfloor$-sized volume, where "$\lfloor \ \rfloor$" represents the operation of returning the ceiling of the input.

2) In part two, the output of 3-D max pooling is put through eight LW units one by one. Note that the eight LW units are divided into four groups, which are shown in different colors in Fig. 1. The first group contains one LW unit whose output has 32 channels. The second consists of two units, with the output of each unit having 64 channels. As with the second group, the third has two 128 channel units. The last group has a single unit and the output of which is 256 channels. Instead of using a pooling layer to reduce the size of features, convolution layers with stride = 2 are adopted in the depthwise convolution layers of the first unit within the last three groups.

3) In the last part, the output volume with size $b \times 256 \times \lfloor (L-7/2^4) \rfloor \times \lfloor (S-2/2^4) \rfloor \times \lfloor (S-2/2^4) \rfloor$ is fed to adaptive average pooling in order to adjust the size of features to a fixed value. In this paper, we use adaptive average pooling to average the features along the space and spectrum. In fact, it is important to employ an adaptive average pooling here. If we just flatten the output of the last unit into a vector as for other conventional structures, we would have to adjust the dimensionality of every fully connected layer for each HSI data set as different HSI data sets have different band numbers. When pretraining a model in one HSI data set and transferring it to other different HSI data sets, this problem will become more severe. In this paper, the output of adaptive average pooling is of size $b \times 256 \times 1 \times 1 \times 1$, for any input size. Finally, we feed the output of the adaptive average pooling into fully connected layers and calculate category scores via the action function log_softmax [47]. For a $C$-dimensional input vector $X = (x_1, x_2, \ldots, x_C)$, the log_softmax formulation can be simplified as

$$\text{log\_softmax}(x_i) = \log \left( e^{x_i} / \sum_{j=1}^{C} e^{x_j} \right). \quad (1)$$

For this paper, $C$ denotes the number of categories.

*2) Training Strategy:* CNNs are learning models, and the kernels of convolution layers and the weight matrix of each fully connected layer in 3-D-LWNet both need to be trained. We take negative log-likelihood as the loss function. In order to avoid overfitting, we add an $L_2$-regularization term to the negative log-likelihood loss to restrict the sum of the squares of the parameters to be small; the formula of which is

$$\text{loss} = \sum_{i=1}^{N} -\text{score}_i^{y_i} + \lambda \|\theta\|_2^2 \quad (2)$$

where $N$ is the number of samples, and $y_i$ is the label of sample $i$. $\lambda$ is weight decay that is used to control the proportion of regularization item in loss function, which is herein empirically set to $1e-5$, for any data sets, and $\theta$ donates all of the parameters of the network. Note that a test of adding a dropout layer, where the probability of an element to be zeroed is set to 0.5, has shown that dropout layer does not work very well in 3-D-LWNet. This is not surprising since adding a dropout layer does not always lead to a positive impact on the improvement of the classification performance. Whether it helps depends on what structure a certain DL network has. He *et al.* [23] also did not use dropout in ResNet. The optimizer we adopt is stochastic gradient descent (SGD) with momentum [10].

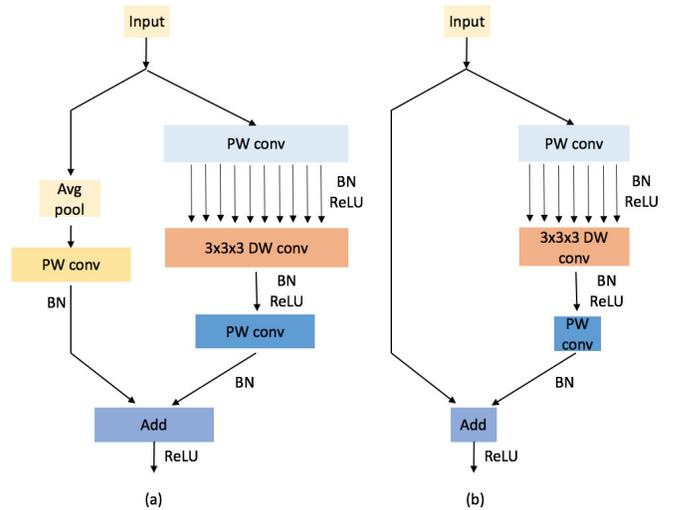

Fig. 2. Lightweight units. (a) Lightweight unit with stride = 2. As for the case where the channel number of the output is doubled and stride = 2, an average pooling layer is added with stride = 2, kernel size = 2, and a pointwise convolution layer is also added onto the shortcut path to reduce the space size and double the channel dimensionality to match the output of the corresponding part. (b) Lightweight unit with pointwise convolution (abbreviated as PW conv) and depthwise convolution (abbreviated as DW conv).

### B. Lightweight Unit

A large number of kernels (parameters) may be prone to overfitting. To alleviate this problem, we take an approach that takes the advantage of depthwise convolution, pointwise convolution, batch normalization, and shortcut connection, resulting in a proposed novel LW unit. The details of this LW unit are shown in Fig. 2. From the top to bottom, the unit contains a pointwise convolution layer (abbreviated as PW conv in the figure), 3-D depthwise convolution layer with $3\times3\times3$-size kernel ($3\times3\times3$ DW conv, for short), and another pointwise convolution layer. Each of the first two convolution layers is followed by a batch normalization layer and a ReLU activation layer, sequentially. After the second pointwise convolution layer, there is only one batch normalization layer. In the add layer, the output features of the right path are added to the features obtained from shortcut connection in an elementwise manner. In the first pointwise convolution layer, the number of channels is increased to $t$ times that of the input channels. The channel number of depthwise convolution layer is the same as that of the first pointwise convolution layer. The parameter $t$ was set empirically. We tested [1/4, 1/2, 2, 4, 6] and empirically chose $t = 4$ as a tradeoff between the performance and parameter size. Assuming the same width, as compared to the depthwise convolution layer, the conventional convolution layer contains many more parameters. Therefore, when we replace the conventional convolution layer with a depthwise convolution layer, it is reasonable to increase the width slightly. Compared with bottleneck, both the parameter size and FLOPs of the LW unit are much less. For instance, in $D$ input channels structure, bottleneck has $32D^2$ parameters, whereas LW unit has $8D^2 + 108D$ parameters. For $S \times S \times L$-sized input, bottleneck requires $32SSLD^2$ FLOPs and LW unit has $SSD(8D^2 + 108D)$ FLOPs.

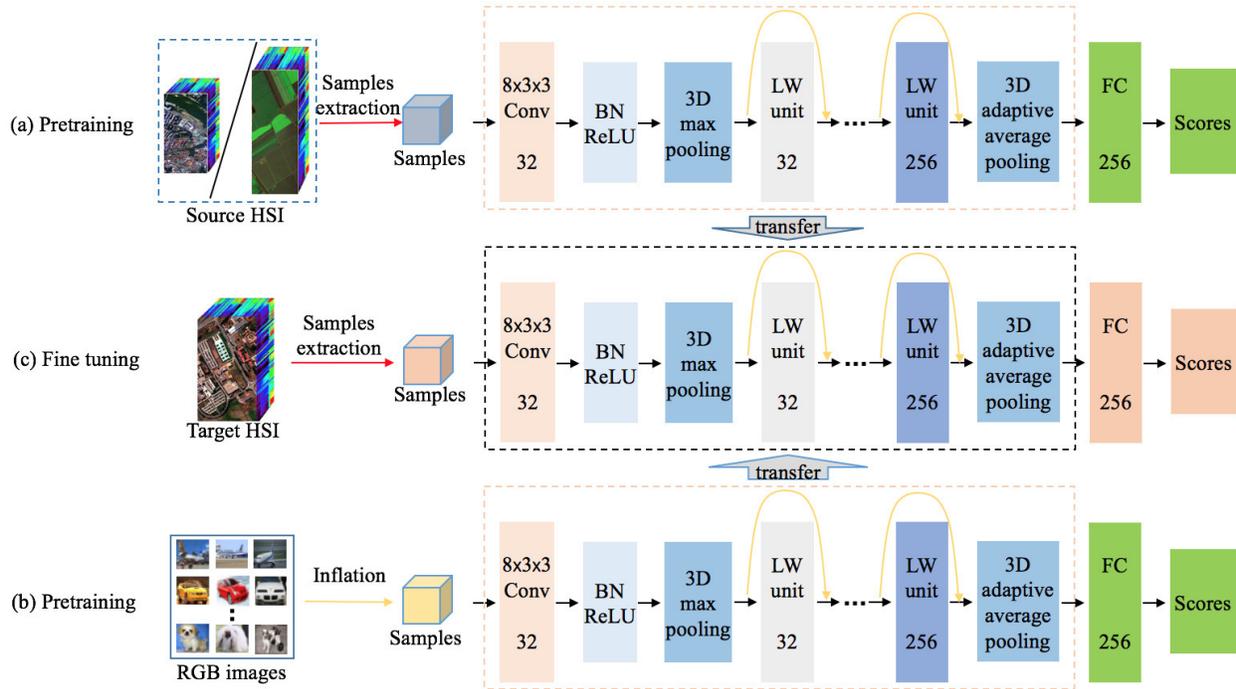

Fig. 3. Transfer learning strategies. (a) Cross-sensor strategy. Pretraining 3-D-LWNet in the source HSI data set that consists of relatively more labeled samples than the target HSI data set. Here, no constraints are imposed in that the source HSI data set has to be collected by the same sensor as the target HSI data sets. (b) Cross-modal strategy. Pretraining 3-D-LWNet in the 2-D RGB image data set. Inflating the 2-D RGB images to 3-D cubes then pretraining the 3-D-LWNet in the inflated data set. (c) Fine-tuning strategy. After 3-D-LWNet is pretrained, the entire model is transferred except fully connected layers to the network built for the target HSI data set as initialization and then the transferred part and the fully connected layer are fine-tuned on the target HSI data set.

There are two types of shortcut connection designed for different situations. Regarding the case where the stride of LW unit is 2 and the channel number of the output is doubled, we add an average pooling layer with stride = 2, kernel size = 2 on the shortcut path to ensure that the output size of the shortcut connection is the same as that of the second pointwise convolution layer of the right path, and add a pointwise convolution layer on the shortcut path to double the width of the output of the shortcut connection. This is different from what bottleneck does. In bottleneck, only a pointwise convolution layer is added with stride = 2 on the shortcut path, which is similar to downsampling and may possibly miss important information. Fig. 2(b) shows another case, where both channel number and size of the output (light blue box) are the same as that of the input (yolk yellow box). In this situation, we add nothing on the shortcut path, just operating on elementwise addition instead.

C. Transfer Learning

In HSI, sample annotation is both time-consuming and resource-consuming, so the number of labeled samples is limited. In this paper, we adopt transfer learning to overcome this problem. The flowchart is shown in Fig. 3.

In Fig. 3, we can see that there are three parts in this system: two pretraining parts and one fine-tuning part. Here, we introduce the transfer learning strategies in detail.

1) *Cross-Sensor Strategy*: Pretraining the model in the source HSI data sets [Fig. 3(a)] consists of relatively more labeled samples than the target HSI data sets and then transferring the pretrained model to the target HSI data sets to fine-tune [Fig. 3(c)]. There are two key points that should be noticed.

   a) The relationship between the source HSI and the target HSI; ideally, we hope that the source HSI data sets and the target HSI data sets are captured by the same hyperspectral remote sensor. For HSIs, this constraint is really onerous. Among the several public HSI data sets, only two pairs of data sets can meet this restriction. In this paper, we implement the proposed approach without such a restriction. To overcome the problem that the data sets captured by different sensors may have different numbers of spectral bands, adaptive average pooling is added to the front of each fully connected layer to adjust the dimensionality of the output features of LW units to a fixed size.

   b) The HSI data sets collected by different sensors may have different spectral configurations. We believe that knowledge learned from different HSI data sets with different spectral configurations can be transferred. This is supported by the observation that knowledge transferring between data sets in different fields has been reported in [48].

2) *Cross-Modal Strategy*: Pretraining the model in a natural RGB image data set [Fig. 3(b)] and then transferring it to target HSIs for fine-tuning. As the strategy that transferring classification models between HSI data sets collected by different sensors has been shown to work, we go one step further here, that is, we transfer models between data sets of different data modalities that exhibit different data characteristics (namely, from natural RGB image modality to HSI modality). Compared with HSI data sets, natural RGB image data sets (e.g., ImageNet, COCO, and CIFAR) have many more labeled samples. The biggest problem for transfer learning strategy two is dimensionality mismatch. There are two different ways to resolve this problem: 1) pretraining a 2-D model in a natural image data set and then inflating the model to 3-D model and 2) inflating the 2-D natural images to 3-D cubes and then pretraining the 3-D model on inflated data sets. In this paper, we adopt the latter for easy implementation. More specifically, we repeat each $m \times n \times$ 3-sized RGB image $l$ times along the third dimension to get the $m \times n \times 3l$-sized cubes. In this paper, $l$ is empirically set to 12.

3) *Fine-Tuning Strategy*: After the model is pretrained via transfer learning strategy one or two, we transfer the entire model except fully connected layers to the network built for the target HSI data set as initialization. Note that the fully connected layers of the network of target HSI are randomly initialized. During fine-tuning, both the transferred part and the randomly initialed part are trained with the same learning rate $\alpha$ by the use of SGD with momentum.

## IV. EXPERIMENTAL RESULTS

### A. Data Description and Experiment Design

In order to evaluate the performance of 3-D-LWNet with transfer learning, we compare it with two other DL-based HSI classification methods: 3-D-CNN-LR [20] and two-CNN [46], on three public HSI data sets: Pavia University, Indian Pines, and Kennedy Space Center (KSC). We also employ two HSI data sets, Salinas and Pavia Center, as source HSI data sets and take two natural RGB image data sets, CIFAR-10 and CIFAR-100, as source data sets to pretrain the models in conducting the transfer learning experiments.

*1) Data Description:* Pavia University and Pavia Center were captured by the reflective optics system imaging spectrometer (ROSIS) sensor in 2001, during a flight campaign over Pavia, northern Italy. Uncorrected data sets contain 115 spectral bands, ranging from 0.43 to 0.86 $\mu$m, with each having a spatial resolution of 1.3 m per pixel. After removing the noisiest data points, Pavia University has 103 bands and Pavia Center has 102 bands. Both Pavia University and Pavia Center are differentiated into nine ground truth classes. The false-color composites of these two data are shown in Fig. 4(a) and (d), respectively.

Indian Pines and Salinas were acquired by the airborne visible/infrared imaging spectrometer (AVIRIS) sensor in 1992. The former was gathered over the Indian Pines test site in North-western Indiana. Uncorrected data contain 224 spectral bands, ranging from 0.4 to 2.5 $\mu$m. It consists of 145×145 pixels with a moderate spatial resolution of 20 m. The number of bands of corrected data is reduced to 200 by removing bands covering the region of water absorption. The next data set was collected over Salinas Valley, California. It has 512×512 pixels and a higher spatial resolution of 3.7 m per pixel. As with Indian Pines, the water absorption bands are also removed in corrected data. The ground truths of them both contain 16 classes.

TABLE I
SAMPLES DISTRIBUTION FOR PAVIA UNIVERSITY

| No. | Class | # of training samples | # of validation samples | # of test samples |
|---|---|---|---|---|
| 1 | Asphalt | 160 | 40 | 6431 |
| 2 | Meadows | 160 | 40 | 18449 |
| 3 | Gravel | 160 | 40 | 1899 |
| 4 | Trees | 160 | 40 | 2899 |
| 5 | Painted metal sheets | 160 | 40 | 1145 |
| 6 | Bare Soil | 160 | 40 | 4829 |
| 7 | Bitumen | 160 | 40 | 1130 |
| 8 | Self-blocking Bricks | 160 | 40 | 3482 |
| 9 | Shadows | 160 | 40 | 747 |
| | Total | 1440 | 360 | 40976 |

The last HSI data set KSC was acquired by the AVIRIS instrument over KSC, Florida, in 1996. It has a spatial resolution of 18 m and wavelength coverage ranging from 0.4 to 2.5 $\mu$m. After removing water absorption and low SNR bands, 176 bands remain for the analysis, with 13 classes representing the various land-cover types were defined for classification.

In transfer learning between RGB natural images and HSIs, we employ two data sets CIFAR-10 and CIFAR-100, each consisting of 60 thousand 32×32 color images. The CIFAR-10 has ten classes, with 6000 images per class. There are 50 000 training images and 10 000 test images. The CIFAR-100 has 100 classes. For each class, there are 500 training images and 100 test images.

*2) Experiment Design:* There are two comparative parts of experiments on classification: those without transfer learning and those with transfer learning.

In experiments of classification without transfer learning, we split each target HSI data set into three subsets, training set, validation set, and test set. The details of distribution are listed in Tables I–III.

In experiments of classification with transfer learning, we randomly extracted 200 samples from each category as test samples and take the rest as the training samples in Pavia Center. Also, 100 labeled samples are randomly extracted from each class for testing in Salinas because the number of labeled samples in Salinas is less than that of Pavia Center. For CIFAR-10 and CIFAR-100, we adopt the default distribution. Thus, the experimental investigations collectively show a diversity of data set usages in training and testing.

### B. 3-D-ResNet

For spectral–spatial classification methods that are based on 3-D-CNN, there are two key points to notice as follows.

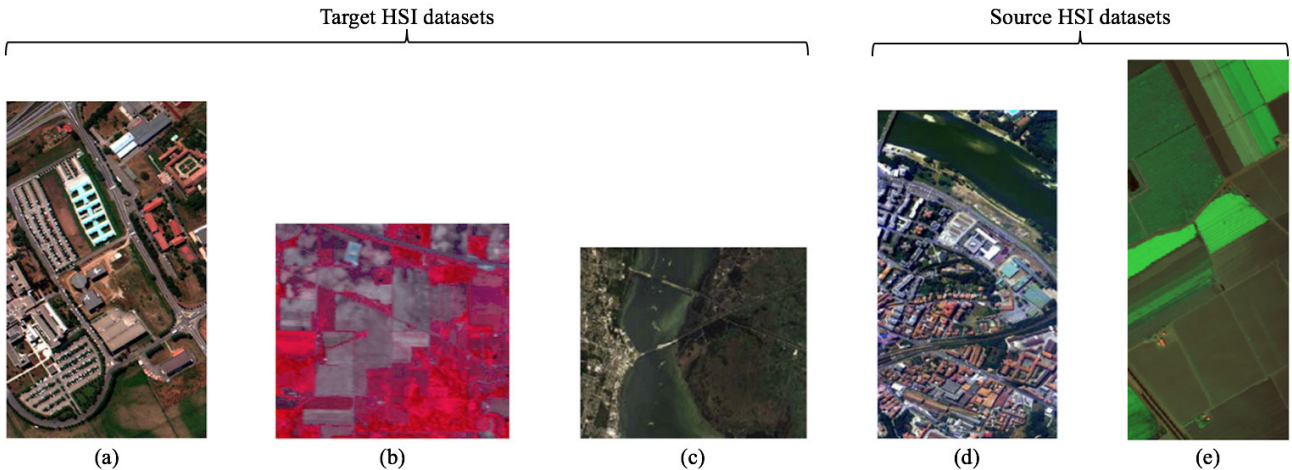

Fig. 4. False color composites of experiment HSI data sets. (a) Pavia University. (b) Indian Pines. (c) KSC. (d) Pavia Center. (e) Salinas.

TABLE II
SAMPLES DISTRIBUTION FOR INDIAN PINES

| No. | Class | # of training samples | # of validation samples | # of test samples |
|---|---|---|---|---|
| 1 | Alfalfa | 24 | 6 | 16 |
| 2 | Corn-notill | 120 | 30 | 1198 |
| 3 | Corn-min | 120 | 30 | 232 |
| 4 | Corn | 80 | 20 | 5 |
| 5 | Grass-pasture | 120 | 30 | 139 |
| 6 | Grass-trees | 120 | 30 | 580 |
| 7 | Grass-pasture-mowed | 16 | 4 | 8 |
| 8 | Hay-windrowed | 120 | 30 | 130 |
| 9 | Oats | 12 | 3 | 5 |
| 10 | Soybean-notill | 120 | 30 | 675 |
| 11 | Soybean-mintill | 120 | 30 | 2032 |
| 12 | Soybean-clean | 120 | 30 | 263 |
| 13 | Wheat | 120 | 30 | 55 |
| 14 | Woods | 120 | 30 | 793 |
| 15 | Buildings-Grass-Tree | 40 | 10 | 49 |
| 16 | Stone-Steel-Towers | 40 | 10 | 43 |
| | Total | 1412 | 353 | 6223 |

TABLE III
SAMPLES DISTRIBUTION FOR KSC

| No. | Class | # of training samples | # of validation samples | # of test samples |
|---|---|---|---|---|
| 1 | Scrub | 27 | 6 | 314 |
| 2 | Willow Swamp | 19 | 4 | 220 |
| 3 | CP Hammock | 19 | 5 | 232 |
| 4 | Slash Pine | 19 | 5 | 228 |
| 5 | Oak/Broadleaf | 12 | 3 | 146 |
| 6 | Hardwood | 18 | 4 | 207 |
| 7 | Swamp | 8 | 1 | 96 |
| 8 | Graminoid Marsh | 31 | 7 | 352 |
| 9 | Spartina Marsh | 41 | 10 | 469 |
| 10 | Cattail Marsh | 31 | 8 | 365 |
| 11 | Salt Marsh | 33 | 8 | 378 |
| 12 | Mud Flats | 39 | 10 | 454 |
| 13 | Water | 73 | 18 | 836 |
| | Total | 370 | 89 | 4297 |

1) The space (window) size of samples. The space size decides on how much contextual information is to be used in each sample cube. A bigger space size implies more contextual information and a smaller space leads to less contextual information. In this paper, we set the space size to $27 \times 27$, following the practice in [20].

2) The structure of 3-D-CNN employed. In general, different network structures mean different classification performances, especially for DL-based methods. Yet, to design an appropriate structure for one specific task is very hard. The depth, width, kernel size, and convolution method all need to be tested. To the best of our knowledge, there is no generic theoretical approach for architecture design. Having observed the outstanding performance of 2-D ResNet, we employ 3-D ResNet to act as the basic structure. Note that He *et al.* [23] proposed five 2-D ResNets: ResNet-18, ResNet-34, ResNet-50, ResNet-101, and ResNet-152. The last two are too deep for HSI in the situation of limited samples, so we employ the first three models and inflate them into corresponding 3-D ResNets: ResNet-18, ResNet-34, and ResNet-50. Finally, we adjust the depth of these three 3-D ResNets and propose seven other 3-D ResNets. Thus, there are ten 3-D ResNet in total, five 3-D ResNets (ResNet-10, ResNet-14_a, ResNet-18, ResNet-34, and ResNet-38) that do not employ bottleneck in their structure and their counterparts that use bottleneck layers. We present the specification of different architectures employed in the experiments in Table IV.

Based on ResNet, we propose ten models as candidates for HSI classification. For selecting the best one from the ten candidate models, we implement fivefold cross validation to estimate these models. The one with the highest average accuracy (AA) in validation sets is taken as the baseline. All of the ten models are applied to three target HSI data sets and trained with the same settings. The results of the ten candidates are listed in Table V and shown in Fig. 6.

TABLE IV
ARCHITECTURES OF 3-D-RESNETS

| basic block | channel number | ResNet-10 | ResNet-14_a | ResNet-18 | ResNet-34 | ResNet-38 |
|---|---|---|---|---|---|---|
| conv | 32 | [8 × 3 × 3], stride=1 | | | | |
| max pool | 32 | [2 × 3 × 3], stride=2 | | | | |
| blocks x1 | 32 | $\begin{bmatrix}3\times3\times3,32\\3\times3\times3,32\end{bmatrix}\times1$ | $\begin{bmatrix}3\times3\times3,32\\3\times3\times3,32\end{bmatrix}\times1$ | $\begin{bmatrix}3\times3\times3,32\\3\times3\times3,32\end{bmatrix}\times2$ | $\begin{bmatrix}3\times3\times3,32\\3\times3\times3,32\end{bmatrix}\times3$ | $\begin{bmatrix}3\times3\times3,32\\3\times3\times3,32\end{bmatrix}\times3$ |
| blocks x2 | 64 | $\begin{bmatrix}3\times3\times3,64\\3\times3\times3,64\end{bmatrix}\times1$ | $\begin{bmatrix}3\times3\times3,64\\3\times3\times3,64\end{bmatrix}\times2$ | $\begin{bmatrix}3\times3\times3,64\\3\times3\times3,64\end{bmatrix}\times2$ | $\begin{bmatrix}3\times3\times3,64\\3\times3\times3,64\end{bmatrix}\times4$ | $\begin{bmatrix}3\times3\times3,64\\3\times3\times3,64\end{bmatrix}\times5$ |
| blocks x3 | 128 | $\begin{bmatrix}3\times3\times3,128\\3\times3\times3,128\end{bmatrix}\times1$ | $\begin{bmatrix}3\times3\times3,128\\3\times3\times3,128\end{bmatrix}\times2$ | $\begin{bmatrix}3\times3\times3,128\\3\times3\times3,128\end{bmatrix}\times2$ | $\begin{bmatrix}3\times3\times3,128\\3\times3\times3,128\end{bmatrix}\times6$ | $\begin{bmatrix}3\times3\times3,128\\3\times3\times3,128\end{bmatrix}\times7$ |
| blocks x4 | 256 98. | $\begin{bmatrix}3\times3\times3,256\\3\times3\times3,256\end{bmatrix}\times1$ | $\begin{bmatrix}3\times3\times3,256\\3\times3\times3,256\end{bmatrix}\times1$ | $\begin{bmatrix}3\times3\times3,256\\3\times3\times3,256\end{bmatrix}\times2$ | $\begin{bmatrix}3\times3\times3,256\\3\times3\times3,256\end{bmatrix}\times3$ | $\begin{bmatrix}3\times3\times3,256\\3\times3\times3,256\end{bmatrix}\times3$ |
| pool | 256 | global average pooling | | | | |
| bottle neck | channel number | ResNet-14_b | ResNet-20 | ResNet-26 | ResNet-50 | ResNet-56 |
| conv | 32 | [8 × 3 × 3], stride=1 | | | | |
| max pool | 32 | [2 × 3 × 3], stride=2 | | | | |
| blocks x1 | 32 | $\begin{bmatrix}1\times1\times1,32\\3\times3\times3,32\\1\times1\times1,128\end{bmatrix}\times1$ | $\begin{bmatrix}1\times1\times1,32\\3\times3\times3,32\\1\times1\times1,128\end{bmatrix}\times1$ | $\begin{bmatrix}1\times1\times1,32\\3\times3\times3,32\\1\times1\times1,128\end{bmatrix}\times2$ | $\begin{bmatrix}1\times1\times1,32\\3\times3\times3,32\\1\times1\times1,128\end{bmatrix}\times3$ | $\begin{bmatrix}1\times1\times1,32\\3\times3\times3,32\\1\times1\times1,128\end{bmatrix}\times3$ |
| blocks x2 | 64 | $\begin{bmatrix}1\times1\times1,64\\3\times3\times3,64\\1\times1\times1,256\end{bmatrix}\times1$ | $\begin{bmatrix}1\times1\times1,64\\3\times3\times3,64\\1\times1\times1,256\end{bmatrix}\times2$ | $\begin{bmatrix}1\times1\times1,64\\3\times3\times3,64\\1\times1\times1,256\end{bmatrix}\times2$ | $\begin{bmatrix}1\times1\times1,64\\3\times3\times3,64\\1\times1\times1,256\end{bmatrix}\times4$ | $\begin{bmatrix}1\times1\times1,64\\3\times3\times3,64\\1\times1\times1,256\end{bmatrix}\times5$ |
| blocks x3 | 128 | $\begin{bmatrix}1\times1\times1,128\\3\times3\times3,128\\1\times1\times1,512\end{bmatrix}\times1$ | $\begin{bmatrix}1\times1\times1,128\\3\times3\times3,128\\1\times1\times1,512\end{bmatrix}\times2$ | $\begin{bmatrix}1\times1\times1,128\\3\times3\times3,128\\1\times1\times1,512\end{bmatrix}\times2$ | $\begin{bmatrix}1\times1\times1,128\\3\times3\times3,128\\1\times1\times1,512\end{bmatrix}\times6$ | $\begin{bmatrix}1\times1\times1,128\\3\times3\times3,128\\1\times1\times1,512\end{bmatrix}\times7$ |
| blocks x4 | 256 | $\begin{bmatrix}1\times1\times1,256\\3\times3\times3,256\\1\times1\times1,1024\end{bmatrix}\times1$ | $\begin{bmatrix}1\times1\times1,256\\3\times3\times3,256\\1\times1\times1,1024\end{bmatrix}\times1$ | $\begin{bmatrix}1\times1\times1,256\\3\times3\times3,256\\1\times1\times1,1024\end{bmatrix}\times2$ | $\begin{bmatrix}1\times1\times1,256\\3\times3\times3,256\\1\times1\times1,1024\end{bmatrix}\times3$ | $\begin{bmatrix}1\times1\times1,256\\3\times3\times3,256\\1\times1\times1,1024\end{bmatrix}\times3$ |
| pool | 1024 | global average pooling | | | | |

In our experiments, we employ SGD as the optimizer and empirically set the initialization learning rate and momentum to 0.01 and 0.9, respectively, which are widely used for training DL models in image classification and action recognition tasks [10], [23], [34], [49], [50]. We train models with different weight decays to find the optimum weight decay from { 1e-3, 1e-4, 1e-5, 1e-6, 1e-7}. Based on the performance in validation sets, we set the weight decay to 1e-5.

Note that a large batch size means taking more GPU memory. As the tested HSI data sets involve 16 categories at most per data set, setting the batch size to 20 is sufficient to perform optimization. The number of training epochs and the learning rate adjustment strategy are empirically decided by analyzing the convergence curve, which is shown in Fig. 5.

In Fig. 5, we can see that the validation loss is saturated after 40–50 epochs for all the three data sets. Thus, we divide the learning rate by 10 at epoch 50 and train the model for ten more epochs. A similar strategy to this is also adopted in [49] and [50].

In conclusion, we employ SGD as our optimizer, where the momentum, weight decay, batch size, the number of training epochs, and the initialization learning rate for SGD optimizer are set to 0.9, 1e-5, 20, 60, and 0.01, respectively. During the final ten epochs, the learning rate is decreased to 0.001.

The results of 3-D ResNets are shown in Fig. 6 and listed in Table V. Compared with models without any bottleneck, those structures that use bottlenecks generally achieve a better performance in terms of classification accuracy, as shown in Fig. 6. These results are close to those in [23]. On the experiments regarding Pavia University and KSC, almost all the networks with bottleneck outperform their corresponding versions that do not use bottlenecks. These experimental results again demonstrate the efficacy of using bottleneck layers. In addition, both structures exhibit a peak in accuracy over the test samples. For instance, in the experiments on the Pavia University scene data set, the classification accuracy over the test set reaches peak in ResNet-20 but starts falling as the depth of network increases. Similarly, among the models that do not employ bottlenecks, ResNet-34 obtains the best performance. As the depth of the network continues to grow, the accuracy drops down.

TABLE V
CROSS-VALIDATION RESULTS OF RESNETS

| Models | Pavia University | | | Indian Pines | | | KSC | | |
|---|---|---|---|---|---|---|---|---|---|
| | OA | AA | K | OA | AA | K | OA | AA | K |
| Basic block | | | | | | | | | |
| ResNet-10 | 97.87±0.11 | 97.87±0.11 | 97.60±0.14 | 98.02±0.02 | 98.54±0.01 | 97.85±0.04 | 92.56±8.27 | 87.91±9.67 | 91.67±10.41 |
| ResNet-14_a | 98.02±0.21 | 98.02±0.21 | 97.78±0.26 | 98.25±0.04 | 98.70±0.03 | 98.06±0.05 | 94.10±8.21 | 90.31±16.49 | 93.40±10.26 |
| ResNet-18 | 97.67±0.20 | 97.67±0.20 | 97.38±0.25 | 98.48±0.03 | 98.89±0.01 | 98.36±0.03 | 94.80±7.91 | 92.19±17.79 | 94.19±9.89 |
| ResNet-34 | 98.01±0.53 | 98.01±0.53 | 97.77±0.66 | 98.30±0.10 | 98.75±0.06 | 98.16±0.12 | 93.40±14.95 | 88.30±18.30 | 92.62±18.70 |
| ResNet-38 | 97.89±0.41 | 97.89±0.41 | 97.63±0.52 | 98.29±0.11 | 98.73±0.07 | 98.15±0.14 | 90.45±14.27 | 86.26±19.25 | 89.33±14.27 |
| Bottleneck | | | | | | | | | |
| ResNet-14_b | 98.24±0.85 | 98.24±0.85 | 98.02±1.12 | 98.44±0.04 | 98.85±0.02 | 98.31±0.05 | 94.80±5.75 | 92.93±18.59 | 94.19±7.17 |
| ResNet-20 | 98.40±0.26 | 98.40±0.26 | 98.20±0.33 | 98.68±0.01 | 99.03±0.06 | 98.57±0.13 | 95.35±3.55 | 93.23±13.52 | 94.79±4.44 |
| ResNet-26 | 98.26±0.39 | 98.26±0.39 | 98.05±0.50 | 98.58±0.16 | 98.96±0.87 | 98.47±0.19 | 95.51±1.08 | 92.84±15.60 | 94.98±1.36 |
| ResNet-50 | 98.01±1.01 | 98.01±1.01 | 97.77±1.28 | 98.31±0.23 | 98.75±0.46 | 98.40±1.13 | 93.96±4.31 | 92.78±7.23 | 93.25±5.39 |
| ResNet-56 | 98.06±0.93 | 98.06±0.93 | 97.81±0.12 | 97.94±0.33 | 98.24±0.54 | 97.63±2.97 | 89.41±7.61 | 83.60±18.59 | 88.12±9.60 |

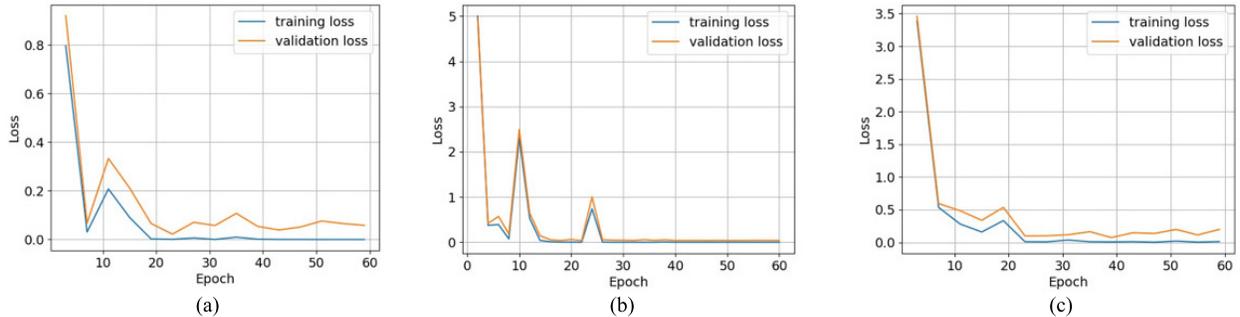

Fig. 5. Convergence curves. (a) Pavia University. (b) Indian Pines. (c) KSC.

Among the three target HSI data sets, ResNet-20 obtains its peak in accuracy on Pavia University and Indian Pines, and ResNet-26 achieves the best performance on KSC. However, on KSC, the gap between the accuracies of ResNet-20 and ResNet-26 is small. The overall accuracy (OA) of ResNet-20 is 95.35%, a mere 0.16 less than that of ResNet-26. Therefore, we choose a unified structure ResNet-20 as the baseline model.

### C. 3-D-LWNet

In this section, we replace the residual unit in the basic model (that employs a unified ResNet-20) with an LW unit for further improvement. The details of the LW unit have been introduced previously (Section III-B). In order to evaluate the performances of 3-D ResNet-20 and 3-D-LWNet, we apply them to three target data sets and compare the classification results with those of running 3-D-CNN-LR [20]. The comparisons are carried out under the same situation as that used in [20] with the training set and test set distributed in the same proportion while running the experiment several times in an effort to achieve statistically averaged performance measures. Tables VI–VIII list the experimental results.

These tables jointly demonstrate the effectiveness of 3-D-LWNet, being capable of obtaining the best performance regarding a range of criteria, including OA, AA, and kappa coefficient ($K$). Here, OA is the ratio between the number of correctly classified samples and that of all samples in the test set, AA is the mean value of the OAs measured over each class, and $K$ is a statistic measurement over the interrate agreement among qualitative items [51]. The 3-D-CNN-LR that is employed adopts the conventional structure and its performance is not so good. 3-D-ResNet employs both shortcut connection and bottleneck, thereby improving the structure and achieving better performance. For HSI classification, however, the structure of 3-D-ResNet is still not sufficiently efficient with room for further improvement. Based on the use of 3-D-ResNet, 3-D-LWNet replaces residual units in 3-D-ResNet with LW units and, hence, reduces the number of parameters involved. From 3-D-CNN-LR to 3-D-ResNet and, then, to 3-D-LWNet, the performance increases step by step. We argue that this is because the structure of the network employed is increasingly more effective. Indeed, on the Pavia University data set, the OA of 3-D-ResNet and 3-D-LWNet is lower than that of 3-D-CNN-LR. Note that 3-D-ResNet and 3-D-LWNet are trained with 1800 training samples, less than half of the training samples required for training 3-D-CNN-LR. Nonetheless, the gap between the OAs of 3-D-CNN-LR and 3-D-LWNet is only 0.14%, with the averaged accuracy of 3-D-LWNet being slightly higher than that of 3-D-CNN-LR. Both 3-D-ResNet and 3-D-LWNet have outperformed 3-D-CNN-LR on Indian Pines and KSC. This is particularly significant for 3-D-LWNet, as it achieves a 1.31% improvement on Indian Pines and 1.91% on KSC.

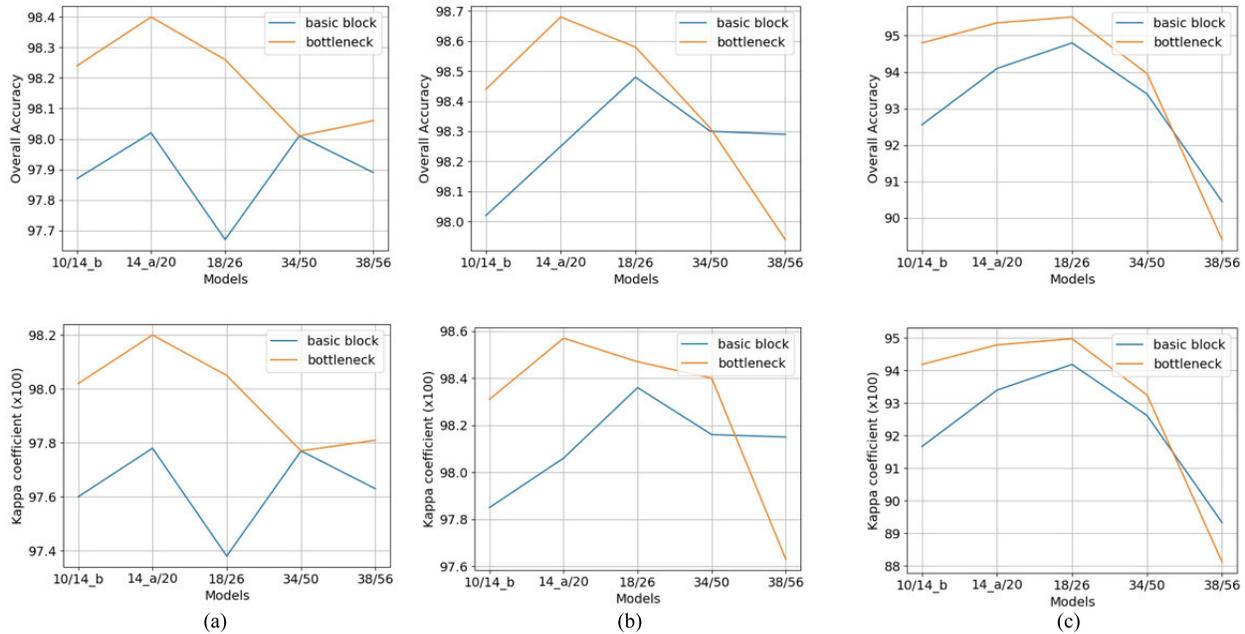

Fig. 6. Classification results of three HSI data sets based on different ResNets. (a) Pavia University. (b) Indian Pines. (c) KSC.

TABLE VI
CLASSIFICATION RESULTS OF PAVIA UNIVERSITY

| models | 2D-CNN-LR | 3D-CNN-LR | 3D ResNet | 3D-LWNet |
|---|---|---|---|---|
| # of training samples | 3930 | 3930 | 1800 | 1800 |
| Asphalt | 97.11±1.04 | 99.36±0.36 | 98.70±1.76 | 98.80±0.25 |
| Meadows | 87.66±1.47 | 99.36±0.17 | 99.32±0.05 | 99.25±0.09 |
| Gravel | 99.69±0.28 | 99.69±0.32 | 98.21±0.79 | 99.87±0.15 |
| Trees | 98.49±0.36 | 99.63±0.15 | 99.36±0.06 | 99.58±0.05 |
| Metal | 100.0±0.00 | 99.95±0.08 | 99.96±0.03 | 100.0±0.00 |
| Bare Soil | 98.00±0.73 | 99.96±0.10 | 99.55±0.32 | 99.95±0.11 |
| Bitumen | 99.89±0.15 | 100.0±0.00 | 99.78±0.04 | 100.0±0.00 |
| Bricks | 99.70±0.27 | 99.65±0.22 | 98.61±1.91 | 99.64±0.12 |
| Shadows | 97.11±1.46 | 99.38±0.61 | 100.0±0.00 | 100.0±0.00 |
| OA | 94.04±0.69 | 99.54±0.11 | 99.18±0.15 | 99.40±0.02 |
| AA | 97.52±0.25 | 99.66±0.11 | 99.28±0.24 | 99.68±0.03 |
| $K$ | 92.43±0.86 | 99.41±0.15 | 98.90±0.27 | 99.20±0.03 |

TABLE VII
CLASSIFICATION RESULTS OF INDIAN PINES

| models | 2D-CNN-LR | 3D-CNN-LR | 3D ResNet | 3D-LWNet |
|---|---|---|---|---|
| # of training samples | 1765 | 1765 | 1765 | 1765 |
| Alfalfa | 99.65±1.47 | 100.0±0.00 | 100.0±0.00 | 100.0±0.00 |
| Corn-notill | 90.64±1.75 | 96.34±1.11 | 97.77±2.37 | 96.61±0.43 |
| Corn-min | 99.11±0.82 | 99.49±0.70 | 99.45±0.12 | 98.87±0.28 |
| Corn | 100.0±0.00 | 100.0±0.00 | 99.82±0.13 | 100.0±0.00 |
| Grass1 | 98.48±1.61 | 99.91±0.23 | 99.25±1.71 | 99.80±0.30 |
| Grass2 | 97.95±2.13 | 99.75±0.30 | 99.96±0.07 | 100.0±0.00 |
| Grass3 | 100.0±0.00 | 100.0±0.00 | 100.0±0.00 | 100.0±0.00 |
| Hay | 100.0±0.00 | 100.0±0.00 | 100.0±0.00 | 100.0±0.00 |
| Oats | 100.0±0.00 | 100.0±0.00 | 100.0±0.00 | 100.0±0.00 |
| Soybean1 | 95.33±2.64 | 98.72±0.95 | 98.48±0.50 | 99.63±1.08 |
| Soybean2 | 78.21±4.93 | 95.52±1.23 | 96.91±1.21 | 98.70±1.02 |
| Soybean3 | 99.39±0.67 | 99.47±0.39 | 99.72±0.12 | 98.87±0.51 |
| Wheat | 100.0±0.00 | 100.0±0.00 | 100.0±0.00 | 100.0±0.00 |
| Woods | 97.71±1.30 | 99.55±0.58 | 100.0±0.00 | 99.91±0.32 |
| Buildings | 99.31±1.40 | 99.54±1.31 | 100.0±0.00 | 99.01±0.30 |
| Stone | 99.22±1.12 | 99.34±1.08 | 100.0±0.00 | 100.0±0.00 |
| OA | 89.99±1.62 | 97.56±0.43 | 98.58±0.12 | 98.87±0.31 |
| AA | 97.19±0.38 | 99.23±0.19 | 99.46±0.38 | 99.45±0.17 |
| $K$ | 87.95±1.90 | 97.02±0.52 | 98.35±0.16 | 98.68±0.15 |

TABLE VIII
CLASSIFICATION RESULTS OF KSC

| models | 2D-CNN-LR | 3D-CNN-LR | 3D ResNet | 3D-LWNet |
|---|---|---|---|---|
| # of training samples | 459 | 459 | 459 | 459 |
| Scrub | 89.43±4.20 | 91.71±3.71 | 93.65±3.59 | 94.87±3.13 |
| Swamp | 87.55±9.13 | 89.73±10.1 | 94.54±5.38 | 95.15±0.27 |
| Hammock | 88.19±7.65 | 92.16±5.33 | 94.74±5.88 | 99.57±0.18 |
| Slash Pine | 81.00±6.68 | 86.94±6.27 | 87.46±10.9 | 94.74±1.34 |
| Oak/Broadl | 96.37±5.18 | 94.79±6.89 | 100.0±0.00 | 97.26±1.41 |
| Hardwood | 84.42±6.78 | 90.92±7.90 | 94.30±0.40 | 97.42±1.47 |
| Swamp | 86.67±9.58 | 91.57±6.14 | 95.83±3.26 | 95.83±4.35 |
| Marsh1 | 96.65±2.49 | 96.22±3.13 | 90.54±1.63 | 97.71±1.82 |
| Marsh2 | 97.05±2.10 | 99.53±0.99 | 100.0±0.00 | 100.0±0.00 |
| Marsh3 | 96.50±2.42 | 99.81±0.37 | 99.84±0.02 | 100.0±0.00 |
| Marsh4 | 94.92±2.95 | 99.79±0.30 | 99.52±0.14 | 99.91±0.02 |
| Mud Flats | 96.94±2.72 | 97.69±2.31 | 97.27±0.57 | 99.78±0.05 |
| Water | 100.0±0.00 | 100.0±0.00 | 100.0±0.00 | 100.0±0.00 |
| OA | 94.11±0.90 | 96.31±1.25 | 96.49±0.58 | 98.22±0.25 |
| AA | 91.98±1.34 | 94.68±1.97 | 95.98±0.54 | 97.87±0.21 |
| $K$ | 93.44±1.00 | 95.90±1.39 | 96.09±0.72 | 98.02±0.30 |

At the first glance, it may seem that the scale of 3-D-LWNet is much larger than that of 3-D-CNN-LR because 3-D-LWNet consists of 20 learnable layers and 3-D-CNN-LR just contains four learnable layers. In fact, benefiting from the depthwise convolution, 3-D-LWNet contains much less parameters than 3-D-CNN-LR. We assume that the 3-D-CNN-LR also employ 3-D convolution layers without bias. Take the network built for Indian pines, for example, the 3-D-CNN-LR has three 3-D convolution layers that contain 44 892 160 parameters in total (the first layer contains $4\times4\times32\times1\times128 = 65\,536$ parameters, the second $5\times5\times32\times128\times192 = 19\,660\,800$ parameters, and the last $4\times4\times32\times192\times256 = 25\,165\,824$ parameters). 3-D-LWNet has a total of 19 convolution layers (the first convolution layer and six LW units, each of which contains three convolution layers), which only involves 763 008 parameters (the first convolution layers have 2304 parameters, the four LW unit groups, respectively, contain 11 648, 71 168, 257 024, and 420 864 parameters), that is, our model saves more than 98.3% compared to the parameters required by the existing work.

### D. Transfer Learning Between Different HSI Data Sets

3-D-LWNet alleviates the problem of limited training samples by reducing the number of parameters and optimizing the structure of the network. It also supports the utilization of transfer learning to provide a good initialization model. In this section, experimental studies are focused on combining 3-D-LWNet with transfer learning.

Note that we do not require that the source HSI data sets and the target HSI data sets have to be captured by the same sensor. This is very different from previous work. Yang et al. [46] also employed transfer learning in their work, but they restricted the data for pretraining to those collected by the same sensor as the target data.

Here, we use five HSI data sets in total. Two data sets, Pavia Center and Salinas, are used for pretraining. Three data sets, Pavia University, Indian Pines, and KSC, are taken as target data sets. Pavia Center and Pavia University were captured by the same sensor ROSIS, both their spatial and spectral resolution are close. Salinas and Indian Pines were collected by AVIRIS, and their spectral resolution is roughly identical. The last target data set KSC was also gathered by AVIRIS, but it only has 176 bands, much less than Salinas. Thus, the basic attributes involved in KSC are rather different from those in the other two source data sets.

Experimental results of transfer learning are listed in Tables IX–XI, where 3-D-LWNet+Pavia and 3-D-LWNet+Salinas represent the models that are pretrained with the Pavia Center data set and the Salinas data set, respectively. In Table IX, we compare the proposed approach with another state-of-the-art transfer learning-based HSI classification method on Pavia University. We randomly chose $\{25, 50, 75\}$ samples from each class of Pavia University for training. From the table, we can obtain the following two conclusions.

1) 3-D-LWNet is better than two-CNN. No matter whether transfer learning is used, the OA of 3-D-LWNet is higher than that of two-CNN (especially when the number of training samples is very small).
2) Transfer learning is useful. When we extract 25 samples per class for training, the OA is improved for 2.76% with transferring from Pavia Center and 4.17% with transferring from Salinas. When we increase the number of training samples to 50 per class, the improvement reduced to 1.36% with transferring from Pavia Center and 2.68% with transferring from Salinas. When the number of training samples is increased to 75 per class, the improvement becomes even smaller. Of course, it can be expected that the improvement provided by transfer learning drops with the increase in training samples. As the number of training samples increases, the model can directly obtain more guidance information from the target HSI data set, so the 3-D-LWNet can work well even without transfer learning.

Note that pretraining in Salinas often achieves better performance than pretraining in Pavia Center. This is also the case regarding Pavia University, which was also collected by the same sensor used for collecting Pavia Center. The experimental results on KSC also illustrate a similar trend. In fact, the models pretrained on Pavia Center and Salinas obtained a similar OA during pretraining and despite the Pavia Center data set contains more labeled samples than Salinas. This implies that transfer learning between homologous data sets (data sets collected by the same sensor) does not necessarily work better than transfer learning between heterologous data sets. The experimental results shown in this section reveal such a seemingly counterintuitive model behavior. This may be because Salinas contains more classes of objects than Pavia Center. In other words, Salinas contains more diverse information. The model pretrained on Salinas has a better generalization ability. In Table X, the experimental results are implemented on Indian Pines and KSC. These results show that transferring from Pavia Center may even harm the performance of 3-D-LWNet on Indian Pines. This may be caused by the fact that Indian Pines contains 16 object classes, whereas Pavia Center only involves 9 object classes. The generalization ability of the model pretrained on Pavia Center is, therefore, not sufficiently powerful to handle the Indian Pines data set.

TABLE IX
TRANSFER LEARNING RESULTS OF PAVIA UNIVERSITY

| # of training samples | without transfer | | with transfer | | | | | |
|---|---|---|---|---|---|---|---|---|
| | Two-CNN | 3D-LWNet | Two-CNN$_4$ | Two-CNN$_3$ | Two-CNN$_2$ | Two-CNN$_1$ | 3D-LWNet +Pavia | 3D-LWNet +Salinas |
| 25 | 68.07 | 88.37 | 74.58 | 77.17 | 77.48 | 77.10 | 91.13 | 92.54 |
| 50 | 79.75 | 95.57 | 84.92 | 85.40 | 85.01 | 85.59 | 96.93 | 98.25 |
| 75 | 85.15 | 97.73 | 87.18 | 87.52 | 86.98 | 86.19 | 98.63 | 99.04 |

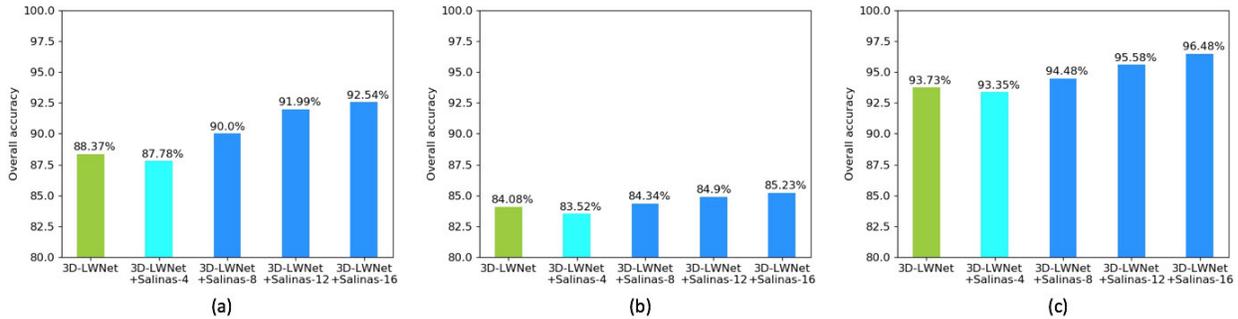

Fig. 7. Transfer learning experiments, 25 training samples per class. (a) Pavia University. (b) Indian Pines. (c) KSC.

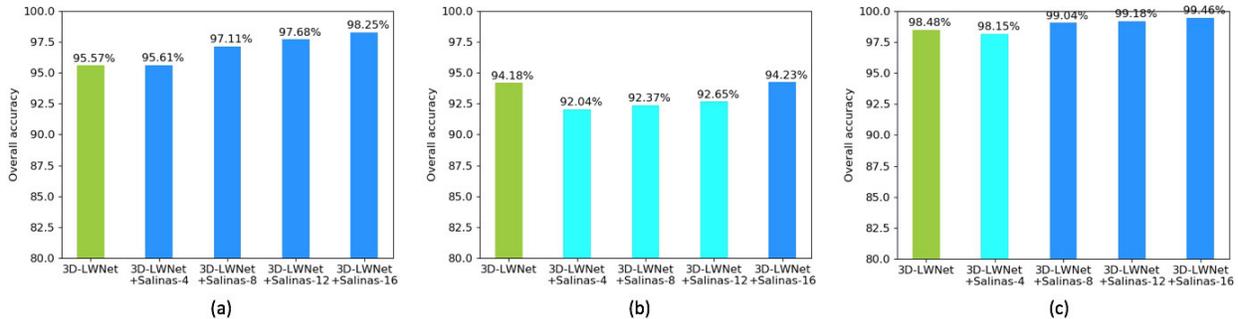

Fig. 8. Transfer learning experiments, 50 training samples per class. (a) Pavia University. (b) Indian Pines. (c) KSC.

### E. Influence of the Number of Object Classes

In Section IV-D, we have found that transferring from Salinas always results in better performance than transferring from Pavia Center. In particular, we hypothesized that this observation might be caused by the fact that Salinas contains more classes of objects. In this section, we extract

TABLE X
TRANSFER LEARNING RESULTS OF INDIAN PINES

| datasets | # of training samples | 3D-LWNet | 3D-LWNet +Pavia | 3D-LWNet +Salinas |
|---|---|---|---|---|
| Indian Pines | 25 | 84.08 | 79.57 | 85.23 |
| | 50 | 94.18 | 92.94 | 94.20 |
| KSC | 25 | 93.73 | 95.06 | 96.48 |
| | 50 | 98.48 | 98.54 | 99.46 |

{4, 8, 12, 16} class subsets from Salinas, pretrain models on these subsets, and then transfer the pretrained models to suit Pavia University, Indian Pines, and KSC for fine-tuning. The experimental results are listed in Table XI and shown in Figs. 7 and 8. The accuracies of 3-D-LWNets are illustrated in yellow–green in each graph, the accuracies of the methods that are better than 3-D-LWNet are shown in dodgerblue, and the accuracies of the methods that are poorer than 3-D-LWNet are shown in cyan.

TABLE XI
TRANSFER LEARNING RESULTS WITH DIFFERENT SUBSETS OF SALINAS DATA SET

| Datasets | # of training samples | 3D-LWNet | 3D-LWNet +Salinas-4 | 3D-LWNet +Salinas-8 | 3D-LWNet +Salinas-12 | 3D-LWNet +Salinas-16 |
|---|---|---|---|---|---|---|
| Pavia University | 25 | 88.37±2.12 | 87.78±5.98 | 90.00±11.89 | 91.99±4.79 | 92.54±11.38 |
|  | 50 | 95.57±0.19 | 95.61±0.74 | 97.11±0.12 | 97.68±0.11 | 98.25±0.06 |
| Indian Pines | 25 | 84.08±4.96 | 83.52±1.12 | 84.34±2.00 | 84.90±2.27 | 85.23±3.84 |
|  | 50 | 94.18±1.10 | 92.04±5.56 | 92.37±4.49 | 92.65±6.05 | 94.23±0.12 |
| KSC | 25 | 93.73±0.71 | 93.35±3.73 | 94.48±0.86 | 95.58±0.93 | 96.48±0.79 |
|  | 50 | 98.48±0.45 | 98.16±2.15 | 99.04±0.15 | 99.18±0.26 | 99.46±0.23 |

In Figs. 7 and 8, we can see that the accuracy of transfer learning grows with the increase in the number of classes of subsets used for pretraining. In addition, when the number of classes of subsets is reduced to 4, pretraining using this subset of data generally causes harm to the resulting classification performance. This is reflected in Fig. 7(a); the accuracy is decreased from 88.37% to 87.78%. Even for Indian Pines (collected by the same sensor as Salinas), pretraining in the subset of Salinas that just involves 4 classes also has a negative influence. The accuracy is reduced by 0.56% when the number of training is set to 25 per class, and 2.14% when this number is set to 50 per class.

*F. Transfer Learning Between RGB Image Data Set and HSI Data Sets*

According to the experimental results of transfer learning as summarized in Sections IV-D and IV-E, we find that transferring between heterologous HSI data sets is feasible and potentially very useful. To further improve this research, in this section, we pretrain a 3-D-LWNet on an inflated RGB images data set prior to fine-tune it on the target HSI data set. Inflation operation has been introduced in Section III-C. Here, we only discuss the experimental results.

Based on the experimental results listed in Table XII, we can draw two main conclusions as follows.

1) Transfer learning between an RGB image data set and an HSI data set works well for HSI classification. It is clear that pretraining in CIFAR-10 and CIFAR-100 improves the classification performance on all three target HSI data sets. Especially for Pavia University, the improvement is significant. When the number of training samples is set to 25 per class, pretraining in Cifar-10 improved the OA by 6.77%, and pretraining in Cifar-100 improved that by 8.58%. Similar to transfer learning between HSI data sets, the improvements provided by pretraining in Cifar-10 and Cifar-100 are drops in the increase of the number of training samples. For example, when the training samples increased to 50 samples per class, the improvements are dropped to 1.71% for Cifar-10 and 2.75% for Cifar-100. Nevertheless, the improvement is still obvious.
2) Diversity of samples used for pretraining has a direct influence on classification accuracy. In the last two sections, the experimental results of transfer learning have shown that the number of object classes within the source HSI data sets plays an important role in transfer learning. The larger such a number, the more improvement in the target data sets. This phenomenon is also shown in transfer learning between RGB images and HSIs. We can see in Table XII that pretraining in Cifar-100 usually achieves a higher classification accuracy than pretraining in Cifar-10.

TABLE XII
EXPERIMENTS RESULTS OF TRANSFER LEARNING BETWEEN RGB IMAGE DATA SET AND HSI DATA SETS

| Datasets | # of training samples | 3D-LWNet | 3D-LWNet + Cifar-10 | 3D-LWNet + Cifar-100 |
|---|---|---|---|---|
| Pavia University | 25 | 88.37±2.12 | 95.14±0.34 | 96.95±0.11 |
|  | 50 | 95.57±0.19 | 97.28±0.18 | 98.32±0.03 |
| Indian Pines | 25 | 84.08±4.96 | 86.31±0.54 | 86.88±1.49 |
|  | 50 | 94.18±1.10 | 92.97±1.06 | 94.11±0.73 |
| KSC | 25 | 93.73±0.71 | 95.55±0.59 | 95.59±1.15 |
|  | 50 | 98.48±0.45 | 99.53±0.22 | 99.56±0.01 |

In addition, an interesting observation is that fine-tuning the model transferred from Cifar-10 can lead to better performance than the model directly trained with Pavia Center and this relationship also exists between Cifar-100 and Salinas. This may also be caused by the diversity of samples. Cifar-100 has 100 classes, which is much larger than the number of classes in Pavia Center or Salinas. Cifar-10 has ten classes, whereas Pavia Center has nine, and it seems that Cifar-10 and Pavia Center are not very different in terms of diversity of samples. In fact, in Cifar-10, even samples belonging to a single class are different. This shows that, in general, the samples of Cifar-10 are more diverse.

V. CONCLUSION

In HSI classification, typically only a limited number of training samples are available. We have addressed this problem with two novel ideas in this paper.

First, we have proposed 3-D-LWNet for spectral–spatial classification of HSIs. Compared to conventional 3-D-CNN that is used for HSI classification, the depth of 3-D-LWNet is much deeper, whereas the number of parameters involved is much less and the classification accuracy of 3-D-LWNet is higher.

Second, we have introduced two transfer learning strategies: cross-sensor strategy and cross-modal strategy and integrated them with 3-D-LWNet for further improvement of the classification performance. With the cross-sensor strategy, we pretrain a 3-D model in the source HSI data set and then transfer the pretrained model to suit the target HSI data set. Unlike the previous work, we not only transfer models between homologous HSI data sets but also do transfer learning between HSI data sets collected by different sensors. Without the restriction that the source HSI data set must be collected by the same sensor as the target HSI data set, the cross-sensor strategy can be applied on more HSI data sets. With the cross-modal strategy, we apply transfer learning on 2-D RGB image data sets and HSI data sets. This is useful as RGB image data sets are generally much larger than HSI data sets, in terms of the amount of labeled samples, and much richer than HSI data sets, in terms of the diversity of samples. The cross-modal strategy builds a bridge between 2-D RGB image data sets and 3-D HSI data sets, which is helpful in taking a full advantage of 2-D RGB image data sets to improve HSI classification performance. To the best of our knowledge, this is the first time that transfer learning between HSI data sets acquired by different sensors and that comparisons between RGB image data sets and HSI data sets have been applied for HSI classification.

CNN is a very powerful machine learning model and has been widely and successfully applied in various fields. However, so far, the design of CNN architectures has been mainly based on experience and empirical experiments. It would, therefore, be very interesting to investigate to optimize the structure of a CNN via intelligent algorithms. This remains an important future work.

REFERENCES

[1] J. Zhu, J. Hu, S. Jia, X. Jia, and Q. Li, "Multiple 3-D feature fusion framework for hyperspectral image classification," *IEEE Trans. Geosci. Remote Sens.*, vol. 56, no. 4, pp. 1873–1886, Apr. 2018.

[2] A. J. Brown *et al.*, "Hydrothermal formation of Clay-Carbonate alteration assemblages in the Nili Fossae region of Mars," *Earth Planetary Sci. Lett.*, vol. 297, nos. 1–2, pp. 174–182, 2010.

[3] A. J. Brown, B. Sutter, and S. Dunagan, "The MARTE VNIR imaging spectrometer experiment: Design and analysis," *Astrobiology*, vol. 8, no. 5, pp. 1001–1011, 2008.

[4] X. Jia, B.-K. Kuo, and M. M. Crawford, "Feature mining for hyperspectral image classification," *Proc. IEEE*, vol. 101, no. 3, pp. 676–697, Mar. 2013.

[5] A. J. Brown, "Spectral curve fitting for automatic hyperspectral data analysis," *IEEE Trans. Geosci. Remote Sens.*, vol. 44, no. 6, pp. 1601–1608, Jun. 2006.

[6] S. Jia, X. Zhang, and Q. Li, "Spectral–spatial hyperspectral image classification using $\ell_{1/2}$ regularized low-rank representation and sparse representation-based graph cuts," *IEEE J. Sel. Topics Appl. Earth Observat. Remote Sens.*, vol. 8, no. 6, pp. 2473–2484, Jun. 2015.

[7] J. A. Benediktsson, J. A. Palmason, and J. R. Sveinsson, "Classification of hyperspectral data from urban areas based on extended morphological profiles," *IEEE Trans. Geosci. Remote Sens.*, vol. 43, no. 3, pp. 480–491, Mar. 2005.

[8] Y. Zhong, A. Ma, and L. Zhang, "An adaptive memetic fuzzy clustering algorithm with spatial information for remote sensing imagery," *IEEE J. Sel. Topics Appl. Earth Observat. Remote Sens.*, vol. 7, no. 4, pp. 1235–1248, Apr. 2014.

[9] Y. Y. Tang, Y. Lu, and H. Yuan, "Hyperspectral image classification based on three-dimensional scattering wavelet transform," *IEEE Trans. Geosci. Remote Sens.*, vol. 53, no. 5, pp. 2467–2480, May 2015.

[10] A. Krizhevsky, I. Sutskever, and G. E. Hinton, "Imagenet classification with deep convolutional neural networks," in *Proc. Adv. Neural Inf. Process. Syst.*, 2012, pp. 1097–1105.

[11] C. Szegedy *et al.*, "Going deeper with convolutions," in *Proc. CVPR*, Jun. 2015, pp. 1–9.

[12] S. Ren, K. He, R. Girshick, and J. Sun, "Faster R-CNN: Towards real-time object detection with region proposal networks," *IEEE Trans. Pattern Anal. Mach. Intell.*, vol. 39, no. 6, pp. 1137–1149, Jun. 2017.

[13] W. Liu *et al.*, "SSD: Single shot multibox detector," in *Proc. Eur. Conf. Comput. Vis. (ECCV)*, Oct. 2016, pp. 21–37.

[14] L. Bertinetto, J. Valmadre, J. F. Henriques, A. Vedaldi, and P. H. S. Torr, "Fully-convolutional siamese networks for object tracking," in *Proc. IEEE Eur. Conf. Comput. Vis. (ECCV)*, Oct. 2016, pp. 850–865.

[15] J. Long, E. Shelhamer, and T. Darrell, "Fully convolutional networks for semantic segmentation," in *Proc. IEEE Conf. Comput. Vis. Pattern Recognit. (CVPR)*, Jun. 2015, pp. 3431–3440.

[16] Y. Chen, Z. Lin, X. Zhao, G. Wang, and Y. Gu, "Deep learning-based classification of hyperspectral data," *IEEE J. Sel. Topics Appl. Earth Observ. Remote Sens.*, vol. 7, no. 6, pp. 2094–2107, Jun. 2014.

[17] Y. Chen, X. Zhao, and X. Jia, "Spectral–spatial classification of hyperspectral data based on deep belief network," *IEEE J. Sel. Topics Appl. Earth Observ. Remote Sens.*, vol. 8, no. 6, pp. 2381–2392, Jun. 2015.

[18] H. Zhang, Y. Li, Y. Zhang, and Q. Shen, "Spectral-spatial classification of hyperspectral imagery using a dual-channel convolutional neural network," *Remote Sens. Lett.*, vol. 8, no. 5, pp. 438–447, 2017.

[19] Y. Li, H. Zhang, and Q. Shen, "Spectral–spatial classification of hyperspectral imagery with 3D convolutional neural network," *Remote Sens.*, vol. 9, no. 1, p. 67, 2017.

[20] Y. Chen, H. Jiang, C. Li, X. Jia, and P. Ghamisi, "Deep feature extraction and classification of hyperspectral images based on convolutional neural networks," *IEEE Trans. Geosci. Remote Sens.*, vol. 54, no. 10, pp. 6232–6251, Oct. 2016.

[21] Z. Zhong, J. Li, Z. Luo, and M. Chapman, "Spectral–spatial residual network for hyperspectral image classification: A 3-D deep learning framework," *IEEE Trans. Geosci. Remote Sens.*, vol. 56, no. 2, pp. 847–858, Feb. 2018.

[22] L. Shu, K. McIsaac, and G. R. Osinski, "Hyperspectral image classification with stacking spectral patches and convolutional neural networks," *IEEE Trans. Geosci. Remote Sens.*, vol. 56, no. 10, pp. 5975–5984, Oct. 2018.

[23] K. He, X. Zhang, S. Ren, and J. Sun, "Deep residual learning for image recognition," in *Proc. IEEE Conf. Comput. Vis. Pattern Recognit. (CVPR)*, Jun. 2016, pp. 770–778.

[24] A. G. Howard *et al.* (2017). "Mobilenets: Efficient convolutional neural networks for mobile vision applications." [Online]. Available: https://arxiv.org/abs/1704.04861

[25] X. Ma, H. Wang, and J. Geng, "Spectral–spatial classification of hyperspectral image based on deep auto-encoder," *IEEE J. Sel. Topics Appl. Earth Observ. Remote Sens.*, vol. 9, no. 9, pp. 4073–4085, Sep. 2016.

[26] W. Hu, Y. Huang, L. Wei, F. Zhang, and H. Li, "Deep convolutional neural networks for hyperspectral image classification," *J. Sensors*, vol. 2015, Jan. 2015, Art. no. 258619.

[27] S. Mei, J. Ji, Q. Bi, J. Hou, Q. Du, and W. Li, "Integrating spectral and spatial information into deep convolutional Neural Networks for hyperspectral classification," in *Proc. Int. Geosci. Remote Sens. Symp. (IGARSS)*, Jul. 2016, pp. 5067–5070.

[28] G. Licciardi, P. R. Marpu, J. Chanussot, and J. A. Benediktsson, "Linear versus nonlinear PCA for the classification of hyperspectral data based on the extended morphological profiles," *IEEE Geosci. Remote Sens. Lett.*, vol. 9, no. 3, pp. 447–451, May 2012.

[29] A. Villa, J. A. Benediktsson, J. Chanussot, and C. Jutten, "Hyperspectral image classification with independent component discriminant analysis," *IEEE Trans. Geosci. Remote Sens.*, vol. 49, no. 12, pp. 4865–4876, Dec. 2011.

[30] K. Makantasis, K. Karantzalos, A. Doulamis, and N. Doulamis, "Deep supervised learning for hyperspectral data classification through convolutional neural networks," in *Proc. IEEE Geosci. Remote Sens. Symp. (IGARSS)*, Jul. 2015, pp. 4959–4962.

[31] J. Yue, W. Zhao, S. Mao, and H. Liu, "Spectral–spatial classification of hyperspectral images using deep convolutional neural networks," *Remote Sens. Lett.*, vol. 6, no. 6, pp. 468–477, Jun. 2015.

[32] L. Mou, P. Ghamisi, and X. X. Zhu, "Deep recurrent neural networks for hyperspectral image classification," *IEEE Trans. Geosci. Remote Sens.*, vol. 55, no. 7, pp. 3639–3655, Jul. 2017.

[33] H. Wu and S. Prasad, "Convolutional recurrent neural networks for hyperspectral data classification," *Remote Sens.*, vol. 9, no. 3, p. 298, 2017.

[34] G. Huang, Z. Liu, L. van der Maaten, and K. Q. Weinberger, "Densely connected convolutional networks," in *Proc. IEEE Conf. Comput. Vis. Pattern Recognit. (CVPR)*, Jul. 2017, pp. 4700–4708.
[35] S. Ioffe and C. Szegedy. (2015). "Batch normalization: Accelerating deep network training by reducing internal covariate shift." [Online]. Available: https://arxiv.org/abs/1502.03167
[36] C. Szegedy, V. Vanhoucke, S. Ioffe, J. Shlens, and Z. Wojna, "Rethinking the inception architecture for computer vision," in *Proc. IEEE Conf. Comput. Vis. Pattern Recognit. (CVPR)*, Jun. 2016, pp. 2818–2826.
[37] C. Szegedy, S. Ioffe, V. Vanhoucke, and A. A. Alemi, "Inception-v4, inception-resnet and the impact of residual connections on learning," in *Proc. AAAI*, vol. 4, 2017, pp. 1–12.
[38] M. Sandler, A. Howard, M. Zhu, A. Zhmoginov, and L.-C. Chen. (2018). "MobileNetV2: Inverted residuals and linear bottlenecks." [Online]. Available: https://arxiv.org/abs/1801.04381
[39] S. J. Pan and Q. Yang, "A survey on transfer learning," *IEEE Trans. Knowl. Data Eng.*, vol. 22, no. 10, pp. 1345–1359, Oct. 2010.
[40] A. Quattoni, M. Collins, and T. Darrell, "Transfer learning for image classification with sparse prototype representations," in *Proc. IEEE Conf. Comput. Vis. Pattern Recognit. (CVPR)*, Jun. 2008, pp. 1–8.
[41] T. Jebara, "Multi-task feature and kernel selection for SVMs," in *Proc. 21st Int. Conf. Mach. Learn.* New York, NY, USA: ACM, 2004, p. 55.
[42] A. Argyriou, T. Evgeniou, and M. Pontil, "Multi-task feature learning," in *Proc. Adv. Neural Inf. Process. Syst.*, 2007, pp. 41–48.
[43] R. K. Ando and T. Zhang, "A framework for learning predictive structures from multiple tasks and unlabeled data," *J. Mach. Learn. Res.*, vol. 6, pp. 1817–1853, Nov. 2005.
[44] R. Raina, A. Y. Ng, and D. Koller, "Constructing informative priors using transfer learning," in *Proc. 23rd Int. Conf. Mach. Learn.*, 2006, pp. 713–720.
[45] A. Torralba, K. P. Murphy, and W. T. Freeman, "Sharing visual features for multiclass and multiview object detection," *IEEE Trans. Pattern Anal. Mach. Intell.*, vol. 29, no. 5, pp. 854–869, May 2007.
[46] J. Yang, Y.-Q. Zhao, and J. C.-W. Chan, "Learning and transferring deep joint spectral–spatial features for hyperspectral classification," *IEEE Trans. Geosci. Remote Sens.*, vol. 55, no. 8, pp. 4729–4742, Aug. 2017.
[47] A. de Brébisson and P. Vincent. (2015). "An exploration of softmax alternatives belonging to the spherical loss family." [Online]. Available: https://arxiv.org/abs/1511.05042
[48] H.-C. Shin *et al.*, "Deep convolutional neural networks for computer-aided detection: CNN architectures, dataset characteristics and transfer learning," *IEEE Trans. Med. Imag.*, vol. 35, no. 5, pp. 1285–1298, 2016.
[49] K. Hara, H. Kataoka, and Y. Satoh, "Can spatiotemporal 3D CNNs retrace the history of 2D CNNs and imagenet?" in *Proc. IEEE/CVF Conf. Comput. Vis. Pattern Recognit.*, Salt Lake City, UT, USA, Jun. 2018, pp. 6546–6555.
[50] J. Carreira and A. Zisserman, "Quo vadis, action recognition? A new model and the kinetics dataset," in *Proc. IEEE Conf. Comput. Vis. Pattern Recognit. (CVPR)*, Jul. 2017, pp. 4724–4733.
[51] W. D. Thompson and S. D. Walter, "A reappraisal of the kappa coefficient," *J. Clin. Epidemiol.*, vol. 41, no. 10, pp. 949–958, Jan. 1988.
[52] R. J. Wang, X. Li, and C. X. Ling, "Pelee: A real-time object detection system on mobile devices," in *Advances in Neural Information Processing Systems*, 2018, pp. 1963–1972.

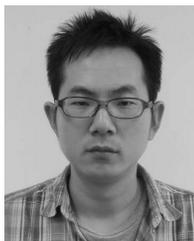
**Haokui Zhang** received the M.S. degree in computer application technology from the Shaanxi Provincial Key Laboratory of Speech and Image Information Processing, Xi'an, China, in 2016. He is currently pursuing the Ph.D. degree with the School of Computer Science, Northwestern Polytechnical University, Xi'an. He is currently pursuing the Joint Ph.D. degree with the School of Computer Science, The University of Adelaide, Adelaide, SA, Australia.

His research interests include cover image processing, deep learning, and hyperspectral image classification.

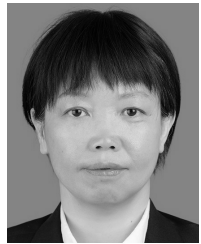
**Ying Li** received the Ph.D. degree in electrical circuit and system from the National Key Laboratory of Radar Signal Processing, Xidian University, Xi'an, China, in 2002.

She is currently a Professor with the School of Computer Science, Northwestern Polytechnical University, Xi'an. Her research interests include image processing, computation intelligence, and signal processing.

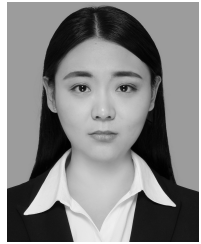
**Yenan Jiang** received the M.S. degree in computer science from Northeast Forestry University, Harbin, China, in 2017. She is currently pursuing the Ph.D. degree with the School of Computer Science, Northwestern Polytechnical University, Xi'an, China.

Her research interests include hyperspectral image processing and deep learning.

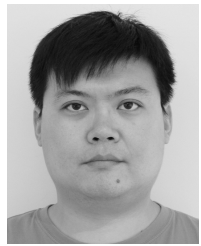
**Peng Wang** received the bachelor's degree in electrical engineering and automation and the Ph.D. degree in control science and engineering from Beihang University, Beijing, China, in 2004 and 2011, respectively.

From 2012 to 2017, he was with the Australian Center for Visual Technologies, The University of Adelaide, Adelaide, SA, Australia. He is currently a Professor with Northwestern Polytechnical University, Xi'an, China.

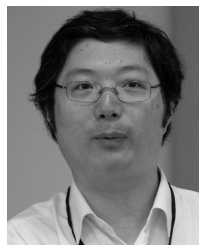
**Qiang Shen** received the Ph.D. degree in computing and electrical engineering from Heriot-Watt University, Edinburgh, U.K., in 1990, and the D.Sc. degree in computational intelligence from Aberystwyth University, Aberystwyth, U.K., in 2013.

He is the Pro Vice-Chancellor for the Faculty of Business and Physical Sciences, Aberystwyth University. He has authored two research monographs and more than 380 peer-reviewed papers.

Dr. Shen was a recipient of an Outstanding Transactions Paper Award from the IEEE.

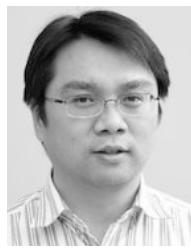
**Chunhua Shen** received the Degree from Nanjing University, Nanjing, China, the Degree from Australian National University, Canberra, ACT, Australia, and the Ph.D. degree from The University of Adelaide, Adelaide, SA, Australia.

He was the Computer Vision Programmer with National ICT Australia, Canberra, ACT. He is currently a Professor of computer science with The University of Adelaide.

Dr. Shen was a recipient of the Australian Research Council Future Fellowship in 2012.